\documentclass{llncs}
\usepackage[utf8]{inputenc}
\usepackage{url}

\begin{document}

\title{3D Segmentation in the Clinic:\\ A Grand Challenge}
\titlerunning{3D Segmentation in the Clinic: A Grand Challenge}
\author{Bram van Ginneken\inst{1} \and Tobias Heimann\inst{2} \and Martin Styner\inst{3}}
\authorrunning{B. van Ginneken, T. Heimann, M. Styner}
\institute{Image Sciences Institute, University Medical Center Utrecht, The Netherlands
\and Medical and Biological Informatics, German Cancer Research Center,\\ Heidelberg, Germany
\and Departments of Psychiatry and Computer Science, University of North Carolina at Chapel Hill, USA}

\maketitle

\begingroup
\renewcommand{\thefootnote}{}
\footnotetext{T. Heimann, M. Styner, B. van Ginneken (Eds.): 3D Segmentation in The Clinic:\\ A Grand Challenge, pp.~7--15, 2007.}
\endgroup

\begin{abstract}
This paper describes the set-up of a segmentation competition for automatic and semi-automatic extraction of the liver from computed tomography scans and the caudate nucleus from brain MRI data. This competition was held in the form of a workshop at the 2007 Medical Image Computing and Computer Assisted Intervention conference. The rationale for organizing the competition is discussed, the training and test data sets for both segmentation tasks are described and the scoring system used to evaluate the segmentation is presented.
\end{abstract}

\section{Introduction}

In 1986 Keith Price wrote a highly amusing article~\cite{price1986} in which he complained that ``computer vision suffers from an overload of written information but a dearth of good evaluations and comparisons.'' He asks the question: ``How do you evaluate the works of others when you do not have their programs? You can look at their results in the papers, but that leaves many questions unanswered.'' And he concludes, after a brief literature survey: ``One area where `non-result' papers are valid (and are lacking) is in a detailed analysis and comparison of various approaches to the same type of problem.'' Although much has changed in two decades and researchers in medical image analysis nowadays have to take evaluation seriously in order to get their work published in high ranking journals, Price's musings still sound all too familiar to many of us.

A wide variety of methods are available for medical image segmentation. Rule-based systems building upon classical image processing techniques remain popular, level-set frameworks are ubiquitous, there is a large variety of methods that learn models of the shape and appearance of objects from training data and can fit these models to unseen data, elastic registration is proving to be a powerful tool for segmentation through atlas matching, classification techniques from pattern recognition and machine learning are attracting a growing number of proponents, and this enumeration is far from complete. Not only is the number of available methods increasing, more and more methods are, or claim to be, generic and applicable to multiple segmentation tasks, usually after applying some suitable modifications and tweaks. Even for experienced researchers in the field it is difficult to choose the appropriate technique for a particular problem.

To facilitate a comparison between a wide range of segmentation techniques we have organized a competition for two important segmentation tasks. The first is liver segmentation from CT scans. This is important for surgery planning, monitoring of therapy, and several other clinical applications. The second is the extraction of the caudate nucleus from brain MRI. The delineation of brain structures such as the caudate is essential for disease understanding, diagnosis, treatment monitoring and functional analysis of the brain. Both tasks are challenging and have been the subject of a range of published papers, but in clinical practice they are routinely segmented completely manually by human experts.

This competition is one of an increasing number of initiatives to achieve scientific progress by sharing data and comparing methods. A survey of such efforts is outside the scope of this paper but we mention a few here. Competitions for liver segmentation and liver cancer extraction have been organized in Japan\footnote{See \url{http://www.tuat.ac.jp/~simizlab/CADM/report_of_competition2005.pdf}}. There exist data repositories for e.g.\ brain\footnote{See \url{http://www.cma.mgh.harvard.edu/ibsr/}} (which supplied data sets for this competition), lung\footnote{See~\cite{armato2004} for CT data and \url{http://www.jsrt.or.jp/cdrom_nodules.html} for chest x-ray data.} and retinal data\footnote{See \url{http://www.isi.uu.nl/Research/Databases/DRIVE}}. Benchmark studies are ongoing for registration\footnote{See \url{http://www.insight-journal.org/rire/}}. Outside the medical domain, there have been competitions for stereo matching\footnote{See \url{http://vision.middlebury.edu/stereo/}}, face recognition\footnote{See \url{http://www.ee.surrey.ac.uk/CVSSP/banca/icpr2004/index.html}}, tracking\footnote{See \url{http://pets2007.net/}} and a 1 million dollar prize for constructing an accurate system to predict movie ratings\footnote{See \url{http://www.netflixprize.com}} is attracting interest from thousands of research teams worldwide.

The competition serves more purposes than a comparative study of a range of algorithms on a common database. It also provides a snapshot of which methods for medical image analysis are currently popular. Moreover, for many researchers it is difficult to obtain a sizeable amount of training and test scans with high quality segmentations such as the one made available for this competition. From the feedback of the participating teams and other researchers who downloaded the data, we conclude that this workshop already sparked many ideas to improve segmentation algorithms.

\section{Data \& organization}

Any team could enter the competition after signing a letter of intent. Both fully automatic and semi-automatic methods were allowed to participate. All papers on caudate segmentation describe fully automatic approaches. For the liver competition, results for both automatic and semi-automatic systems were submitted. In our initial call for papers we stated that user interaction for semi-automatic methods should be limited to pre-processing and initialization of an automatic computation stage. For the liver competition, two papers were submitted that use interaction until the user is completely satisfied with the segmentation result. These papers have also been included in these proceedings, under the heading `Interactive Liver Segmentation' because we believe such an approach may be very attractive for clinical application, and the amount and nature of the interaction among semi-automatic methods varies substantially so it is hard to define a clear boundary between semi-automatic and interactive.

For both competitions three data sets were constructed. The first is a training data set that includes both images and binary masks of the segmentations of the structures of interest, produced by human experts. Participants can use these images to train their algorithm, but they are free to use their own training data in addition to the supplied data. They may also use the training data to determine suitable values for free parameters in their algorithms. The second data set is a test set that was distributed with the training data (both could be downloaded from a web site). The test set did not include segmentations and participants had to send in the segmented test sets before a given deadline.

Finally there is a second test set for both tasks that will be released at the start of the one-day workshop on October 29, 2007. This set is the online test set. Segmentation results are to be submitted within three hours after the online test set has been made available. Participants unable to meet these demands were allowed to make an executable version of their program available to the workshop organizers who processed the data with the supplied software in advance. Completely automatic methods were eligible for cash prizes made available by the sponsors of the workshop. Teams containing members of the workshop organizers were excluded from this on-line competition.

The reason for using two different test sets is that distributing the test data to be segmented together with the training data allows teams to optimize their algorithms for the provided test cases. This can be the case even if the segmentations of the test cases are unknown, and can happen even unintentionally\footnote{Too many of us have learned this the hard way, and some have not learned it yet. From page 483 of~\cite{duda2001} we quote: ``It is essential that the validation (or the test) set not include points used for training the parameters in the classifier --- a methodological error known as `testing on the training set'. A related but less obvious problem arises when a classifier undergoes a long series of refinements guided by the results of repeated testing on the same test data. This form of `training on the test data' often escapes attention until new test samples are obtained.''}. A contest during the actual meeting on new test data also increased the competitive element of the workshop.

All submitted segmentations were processed with a fixed evaluation protocol described in the next section. The following subsections describe the data for both tasks in more detail.

\subsection{Liver data}

All CT images were enhanced with contrast agent and scanned in the central venous phase on a variety of scanners (different manufacturers, 4, 16 and 64 detector rows). As it is CT, all data sets were acquired in transversal direction. The pixel spacing varied between 0.55 and 0.80\,mm, the inter-slice distance varied from 1 to 3\,mm. There was no overlap between neighboring slices.

A total of 40 images was used for the workshop. Most of them were pathologic and included tumors, metastasis and cysts in different sizes. Images have been randomized into three groups: 20 training images, 10 images for the first test set and 10 for the on-line contest.

All segmentations were created manually by radiological experts, working slice-by-slice in transversal view. The first tool they employed was an intensity-based region grower. In case of leakage, these leaks were removed by drawing manual cut-lines. The segmentation was defined as the entire liver tissue including all internal structures like vessel systems, tumors etc. In general, a vessel counts as internal if it is completely surrounded by liver tissue (in the transversal view). For the large vessels that enter the liver (vena cava and portal vein) the parts enclosed by liver tissue, i.e.\ as the convex hull of the liver shape in that area, were included in the segmentation.

\subsection{Caudate data}

The caudate data consisted of a total of 33 training images from two sources: a) 18 datasets of healthy controls from the Internet Brain Segmentation Repository provided by the Center for Morphometric Analysis at Massachusetts General Hospital\footnote{Online available at \url{http://www.cma.mgh.harvard.edu/ibsr}}, b) 15 datasets of healthy controls and subjects in a Schizotypal Personality Disorder (SPD) study~\cite{levitt2002} provided by Psychiatry Neuroimaging Laboratory at the Brigham and Women's Hospital, Boston. The definition of the caudate in these two datasets is not exactly the same. Caudate manual segmentation protocols often differ in regard to the exact separation of caudate to the nucleus accumbens anteriorly and the level of inclusion of the vanishing tail posteriorly. Such variability in caudate definition is commonplace and part of clinical routine when comparing studies. The authors were instructed to state in their paper what training population their method was built upon.

The testing data stems from a variety of sources. The diversity of the test data evaluates the algorithms in respect to a) flexibility to pathology, age group, and signal-to-noise aspects, b) stability in a test/re-test situation, and c) stability in respect to slight differences in caudate definition:

\begin{enumerate}
\item As the first set of data an additional 14 cases were selected from the same SPD study~\cite{levitt2002} provided by Psychiatry Neuroimaging Laboratory at the Brigham and Women's Hospital, Boston. These scans can be considered routine adult data sets.
\item The second and third data sets consist of 5 pediatric (age 2--4 years) and 5 elderly (age 60--75 years) subjects provided by the UNC autism research group and the UNC Parkinson research group, University of North Carolina at Chapel Hill. Both datasets serve to test the versatility of the segmentation methods. The manual segmentation protocol for the caudates was the same for both datasets but slightly different from the training datasets.
\item The last 10 data sets expose the algorithms to a test/re-test situation in order to assess the reliability and stability of the algorithm. A single female subject (age 25 years) was scanned two times within a 24 hour time window each at five different MR sites over a period of six weeks. The age of the subject plus the absence of physical or mental illness suggests that the brain of the subject did not change in the six week period. Four of the five sites acquired the images using a GE Signa 1.5 Tesla scanner and one site was using a Philips Gyro Scan NT 1.5 Tesla scanner. The datasets were provided by the Duke Image Analysis Laboratory (DIAL)~\cite{styner2002}. This dataset allows for an evaluation of reproducibility of image analysis methods. The main sources of variance in this dataset are patient positioning, scanner geometry, scanner intensity variation (RF coil) and discrete image artifacts, e.g.\ motion artifacts.
\end{enumerate}

All datasets were scanned with an Inversion Recovery Prepped Spoiled Grass sequence on a variety of scanners (GE, Siemens, Philips, all 1.5 Tesla). Some data sets were acquired in axial direction, others in coronal direction. All data sets were re-oriented to axial orientation, but were not aligned in any fashion. The resolution of the datasets is at $0.9375 \times 0.9375 \times 1.5$\,mm. All datasets were randomized and authors were blind to the randomization.

\section{Evaluation}

The quality of a segmentation can be evaluated in many different ways. A sensible evaluation criterion depends on the purpose of the segmentation procedure. If the goal is to estimate the volume of an object, the volumetric error would be an obvious criterion. But a segmentation with exactly the same volume as the reference (we refer to the manual segmentations that we consider to be the `truth' as reference) can be completely wrong if a voxel by voxel comparison is made. As a result, there are different evaluation metrics in common use and most papers on segmentation report results in terms of more than one metric. An overview of various metrics and their peculiarities is given in~\cite{gerig2001} and~\cite{niessen2000}.

As we are organizing a competition, we need a single evaluation criterion, but several popular metrics exist. To deal with this situation we devised a scoring system that combines several metrics into a single overall score. To be able to give a meaningful interpretation to these scores, we decided to gauge each score by relating it to the result that could be expected if an independent human observer would perform the segmentation manually. We also wanted to avoid that a few completely failed segmentations would damage an overall score irreparably. Thus we decided to award 100 points for a perfect result (the best value that could be obtained for a metric) and a predefined amount $X$ for a score that is typical for an independent human observer. The scaling between these two gauge values is linear, but negative scores are impossible, so 0 points is the minimum possible value to achieve. The free value of $X$ was chosen such that segmentations that visually corresponded to the subjective notion of a `complete failure' would get 0 points. Some preliminary tests prompted us to set $X = 75$ for the liver and $X = 90$ for the caudate.

The following five metrics were used, formulated in such a way that for each metric, 0 corresponds to a perfect result:

\begin{itemize}
\item Volumetric overlap error, in percent. This is the number of voxels in the intersection of segmentation and reference divided by the number of voxels in the union of segmentation and reference, subtracted from 1 and multiplied by 100. This value is 0 for a perfect segmentation and has 100 as the lowest possible value, when there is no overlap at all between segmentation and reference.
\item Relative absolute volume difference, in percent. The total volume difference of the segmentation to the reference is divided by the total volume of the reference. The result is multiplied by 100. This signed number is reported in the tables of the papers in these proceedings, so one can recognize undersegmentations by negative values and oversegmentation by positive values. To compute a score, the absolute value is taken. Note that the perfect value of 0 can also be obtained for a non-perfect segmentation, as long as the volume of that segmentation is equal to the volume of the reference.
\item Average symmetric surface distance, in millimeters. The border voxels of segmentation and reference are determined. These are defined as those voxels in the object that have at least one neighbor (of their 18 nearest neighbors) that does not belong to the object. For each voxel along one border, the closest voxel along the other border is determined (using Euclidean distance, not signed, and real world distances, so taking into account the different resolutions in the different scan directions). All these distances are stored, for border voxels from both reference and segmentation. The average of all these distances gives the average symmetric absolute surface distance. This value is 0 for a perfect segmentation.
\item Root Mean Square (RMS) symmetric surface distance, in millimeters. This measure is similar to the previous measure, but stores the squared distances between the two sets of border voxels. After averaging the squared values, the root is extracted to give the symmetric RMS surface distance. This value is 0 for a perfect segmentation.
\item Maximum symmetric surface distance, in millimeters. This measure is similar to the previous two, but in this case the maximum of all voxel distances is taken instead of the average. This value is 0 for a perfect segmentation.
\end{itemize}

The reference values for an independent human observer, that yield scores of 75 and 90 for the liver and caudate tasks are listed in Table~\ref{tab:reference}. For the caudate segmentation task, scores were computed independently for the left and right caudate and averaged to get the score per case. For each case, each metric was converted to a score. The overall score for a case was simply the average of the individual metric scores. Tables in the papers in these proceedings list all scores for each test case.

\begin{table}
\centering
\begin{tabular}{lrr}
 & Liver & Caudate \\
Volumetric overlap error [\%] & 6.4 & 15.8 \\
Relative absolute volume difference [\%] & 4.7 & 5.6 \\
Average symmetric surface distance [mm] & 1.0 & 0.27 \\
RMS symmetric surface distance [mm] & 1.8 & 0.56 \\
Maximum symmetric surface distance [mm] & 19 & 3.4 \\
\end{tabular}
\vspace{2mm}
\caption{Reference values for the scoring systems for liver and caudate segmentation.}
\label{tab:reference}
\end{table}

Using this scoring system one can loosely say that 75/90 points for a liver or caudate segmentation is `comparable to human performance'. But this is only a rough indication, as the scores of humans will vary across cases, and across humans. For difficult cases, a score below 75/90 points may therefore still be very good. We have not investigated this in detail for the data sets used in this competition. The reference values for the liver segmentation were computed using a segmentation of the ten test sets by a medical student with no previous experience in liver segmentation and no experience with the segmentation protocol employed by the experienced radiologists who determined the reference. An accurate (interactive) segmentation may therefore well achieve scores above 75 points. To gauge the caudate scoring system, multiple human segmentations on a subset of the test cases (the UNC cases only) were used. They were all performed by experienced operators who were trained to use the segmentation protocol and software. This may make it harder to achieve 90 points for the caudate than to achieve 75 points for the liver.

For the caudate segmentation task, one group of test cases are 10 scans of the same subject acquired on different scanners. For these cases no reference segmentations were available, and these cases therefore do not contribute to the total score. These cases are employed to test if a method is reproducible in a test/retest situation, as well as in presence of scanner variability. This reproducibility or stability measurement is based on the coefficient of variation ($\mathrm{COV} = \frac{\sigma}{\mu}$ times 100\%) of the volumetric measurements. The reference caudate segmentations of human experts show on average a COV of 3.1\%.

In addition to the measures listed above, Pearson correlation between reference and segmentation volumes is also reported for the caudate results. This coefficient captures how well the segmentation volumes correlate with those of the reference in a range of $[-1,1]$. This measure is of high significance in neuro-morphological studies that are still strongly rooted in volumetric assessment, but was not included in the score computation. In such neuro-morphological studies, a systematic bias in the segmentation method that results in clear errors in the above scores, but still very good correlation with the reference is often acceptable. Furthermore, the correlation values should be close to those between human raters, otherwise this indicates that the use of the segmentation method would be a source of additional variation. Using the same expert segmentations as for computing the score values, the average Pearson correlation across human expert segmentations of the caudate is 0.71.

The scores were computed by the workshop organizers after participating teams had uploaded results to a website. Each team received a table with the scores, and a panel with their segmentation and the reference shown on some representative slices. Each paper in the proceedings contains one or more of these tables and panels.

\section{Outlook}

Despite the short period between the call for papers for the workshop was released and the letters of intent for participation had to be sent in, an unexpectedly large number of 44 teams registered and downloaded the data for one or both tasks. Of these, 25 were from Europe, 15 from North America, 3 from Asia and 1 from Australia. More than half of all registered teams submitted results. For the caudate competition 9 papers were eventually accepted for the workshop and are included in these proceedings. For the liver competition 16 papers were accepted. We believe this large response shows that there is a definite interest within the medical image segmentation research community to participate in comparative studies such as these.

We intend to make the data sets and the results of the various systems described in these proceedings publicly available after the workshop. We hope to report on the results of the workshop, including those of the on-line competition, in future publications.

\section{Acknowledgements}

We would like to thank the following people and institutes for making data available for this competition: Andrew Worth and David Kennedy at the Center for Morphometric Analysis at Massachusetts General Hospital; Martha Shenton and James Levitt at the Psychiatry Neuroimaging Laboratory at the Brigham and Women's Hospital, Boston; Joseph Piven and Heather Hazlett Cody at the Autism Research Center at the University of North Carolina; Xuemei Huang at the Movement Disorders Division, Department of Neurology at the University of North Carolina, Cecil Charles at the Duke Image Analysis Laboratory; Christoph Becker at the Department of Clinical Radiology, University Hospital of Munich, Germany; Lars Grenacher at the Department of Diagnostic Radiology, University Hospital of Heidelberg, Germany; Andrés Córdova B.\ at the Department of Oncology, Clínica Alemana de Santiago, Chile. The following grant support is noted: UNC Neurodevelopmental Disorders Research Center HD 03110, NIH RO1 MH61696 and NIMH MH64580, NIH Roadmap for Medical Research Grant U54 EB005149-01, NIH R01 MH50747, and K05 MH070047. We acknowledge the financial support from the workshop sponsors, Siemens Medical Solutions and Chili Radiology. Finally we would like to thank all participants to this workshop for their great efforts and their courage to participate in this contest.

\end{document}